\documentclass[10pt,twocolumn,letterpaper]{article}
\usepackage{cvpr}              

\usepackage{graphicx}
\usepackage{amsmath}
\usepackage{amssymb}
\usepackage{booktabs}
\usepackage[accsupp]{axessibility}

\usepackage[pagebackref,breaklinks,colorlinks]{hyperref}
\usepackage{multirow}
\usepackage{booktabs}
\usepackage[capitalize]{cleveref}
\crefname{section}{Sec.}{Secs.}
\Crefname{section}{Section}{Sections}
\Crefname{table}{Table}{Tables}
\crefname{table}{Tab.}{Tabs.}

\usepackage{comment}

\usepackage[titletoc]{appendix}
\usepackage{lipsum}
\usepackage{multicol}

\def\cvprPaperID{6} 
\def\confName{CVPR}
\def\confYear{2023}

\begin{document}

\title{Sign Language Translation from Instructional Videos}
\author{Laia Tarrés$^{1,2}$ \hspace{0.2cm}
 Gerard I. Gállego$^{1}$ \hspace{0.2cm}
 Amanda Duarte$^{2}$ \hspace{0.2cm}
 Jordi Torres$^{1,2}$ \hspace{0.2cm}
 Xavier Gir{\'o}-i-Nieto$^{3,}$\thanks{ Work done outside of Amazon}
\\
$^1$\emph{Universitat Polit\`{e}cnica de Catalunya} \quad $^2$\emph{Barcelona Supercomputing Center}
\quad $^3$\emph{Amazon}
\\ 
\href{https://imatge-upc.github.io/slt_how2sign_wicv2023}{\small\url{{https://imatge-upc.github.io/slt_how2sign_wicv2023}}}
}
\maketitle

\begin{abstract}

The advances in automatic sign language translation (SLT) to spoken languages have been mostly benchmarked with datasets of limited size and restricted domains.
Our work advances the state of the art by providing the first baseline results on How2Sign, a large and broad dataset.

We train a Transformer over I3D video features, using the reduced BLEU as a reference metric for validation, instead of the widely used BLEU score.
We report a result of 8.03 on the BLEU score, and publish the first open-source implementation of its kind to promote further advances.

\end{abstract}

\section{Introduction}
\label{sec:intro}

Sign language translation (SLT) is the task of translating continuous sign language videos into spoken language sentences.
SLT is a challenging multimodal problem that requires both a precise understanding of the signer's pose and the generation of a textual transcription.
The current state of the art for automatic SLT is still far away from considering the problem solved~\cite{Chen-2022-twostream-SLR-SLT, voskou2021stochastic, de2021frozen, camgoz2020sign_language_transformers, Zhou2021ImprovingSLT-with-monolingual-CSLDaily, yin-read-2020-better-sign-language-translation}. 
Solving SLT will bring important benefits to the communication between signers and non-signers.

Recent advances in SLT have followed a trajectory similar to other computer vision and natural language processing problems: training deep neural networks on large-scale datasets.
However, the availability of public sign language datasets is limited and especially reduced when considering parallel corpus of videos and their textual translations, which allow benchmarking the state of the art.
Up to date, the most used dataset to assess the progress in SLT is PHOENIX-2014-T~\cite{RWTH-PHOENIX-Weather-2014}, with only 9.2 hours of video recordings on the restricted domain of weather forecasts.

\begin{figure}[ht]
\centering
\includegraphics[width=1\linewidth]{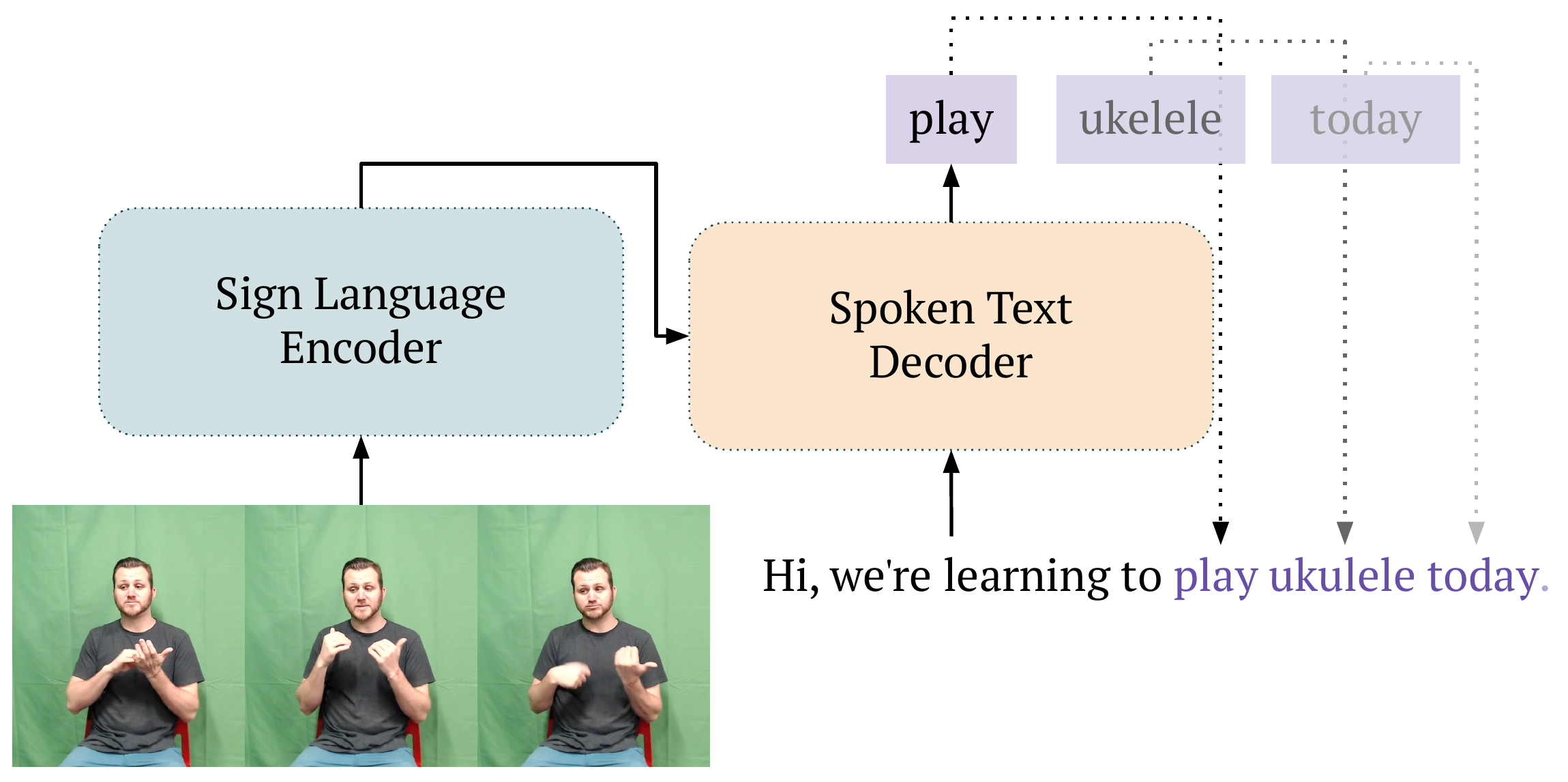}
\caption{A basic pipeline for sign language translation.}
\label{fig:teaser}
\end{figure}
In this work, we consider a much larger and more complex dataset, How2Sign~\cite{Duarte_2021_how2sign}, which contains almost 80 hours of instructional videos from 10 different topics.
This dataset was approved by the Carnegie Mellon University Institutional Review Board. We publish the first SLT baselines for this dataset, achieving a BLEU score of 8.03.

In addition, we show that the common practice of choosing the best model checkpoint based on the BLEU metric may be misleading. This is because the model tends to capture frequent patterns, and may even provide realistic outputs completely unrelated to the input video.
We empirically test how using an alternative metric~\cite{microsoft-wmt-slt22}, \textit{reduced BLEU} (rBLEU), we can better characterize the performance of the SLT solutions and choose better checkpoints during training.

We provide open code and models of a translation system from American Sign Language (ASL) to written English, trained on How2Sign. Our implementation contains scripts to preprocess the data, train, translate, and evaluate models, which allows reproducibility and adaptation to other datasets.\footnote{\href{https://github.com/imatge-upc/slt_how2sign_wicv2023}{https://github.com/imatge-upc/slt{\_}how2sign{\_}wicv2023}}
\section{Related Work}
\label{sec:relatedwork}
Sign language video understanding has been addressed from a variety of tasks: sign language recognition (SLR) over isolated or continuous signs~\cite{Dist_for_CSLR, SkeletonSLR, RGB_SLR, SLR_coArt, SLR_VizAlign, SLR_CrossModAlign, SLR_stochastic}, sign language translation (SLT) \cite{RWTH-PHOENIX-Weather-2014, SLT, SLT_MultiChannel, SLT_Korean}, sign language production (SLP) \cite{SLP_Photo, SLP_GraphSAtt, SLP_Mixed, SLP_ProgTransf, SLT_EvSignNow} or retrieval~\cite{duarte2021signretrieval}.
Our work focuses on sign language translation.

\vspace{0.5em}
\noindent \textbf{Gloss supervision}.
Gloss-based SLT~\cite{2018_Neural_Sign_Language_Translation, camgoz2020sign_language_transformers, Zhou2021ImprovingSLT-with-monolingual-CSLDaily, Chen-2022-twostream-SLR-SLT} uses an intermediate textual representation between the input video sequence and the output text. These tokens are named \textit{glosses}. Glosses are a type of transcription of sign languages that must be produced by trained sign language linguists and that are available in some SLT datasets.
Glosses provide supervision that helps models in their training, but their acquisition is also very time-consuming and expensive because of the scarcity of annotators.

On the other hand, gloss-free SLT~\cite{SLT_MultiChannel, Orbay-2020-SLT-learning-tokenization, Camgoz-2021-content4all, yin-read-2020-better-sign-language-translation, shi-etal-2022-openASL} addresses the raw task of converting the video into text, without any intermediate gloss. 
Our work targets this second case, as glosses for \cite{Duarte_2021_how2sign} have yet to be released.

\vspace{0.5em}
\noindent \textbf{Datasets}.
Gloss-free SLT has traditionally focused on datasets  that have a limited amount of data, and a restricted vocabulary~\cite{AUTSL, WLASL, SLT_Korean, Phoenix, 2018_Neural_Sign_Language_Translation}. 
Thus, the challenge to serve in real-world use cases remains. 

SLT is by definition associated with \textit{continuous} signing rather than \textit{isolated} signing, because signers naturally concatenate one sign after the other with no resting position, similarly to how speakers concatenate spoken words. Compared to isolated SL, continuous SL videos include important effects such as prosody, which can play a crucial role in the meaning of a sentence. 

Table~\ref{tab:SLT_datasets} shows the current state of the art in terms of the BLEU metric for different SLT benchmarks.
Reasonable scores in the range between 29 and 60 BLEU have been reported in three datasets of limited vocabulary size: KETI~\cite{Ko-2019-SLT-based-human-keypoint-estimation},  PHOENIX-2014T~\cite{2018_Neural_Sign_Language_Translation}, and CSL Daily~\cite{Zhou2021ImprovingSLT-with-monolingual-CSLDaily}. 

Our work aims at the more open domain of instructional videos across 10 different topics, to set the first SLT baselines on the How2Sign~\cite{Duarte_2021_how2sign} dataset. This dataset has been used in the past for human motion transfer, sign language retrieval \cite{duarte2021signretrieval}, or topic detection~\cite{budria2022topic}, but never for SLT.

\begin{table*}[!ht]
\centering
\begin{tabular}{lcccccccl}
\toprule
Dataset                                                                               & \multicolumn{3}{c}{Duration(h)} & \multicolumn{3}{c}{Vocabulary(K)}                                              & BLEU                                                                   & Domain                 \\
                                                                                      & train     & val      & test     & \multicolumn{1}{l}{train} & \multicolumn{1}{l}{val} & \multicolumn{1}{l}{test} &                                                                                  &                        \\ \midrule
KETI~\cite{Ko-2019-SLT-based-human-keypoint-estimation}                              & 20.05     & 2.24     & 5.70     & $\leftarrow$ & 0.49 & $\rightarrow$                                                       & 57.37~\cite{Ko-2019-SLT-based-human-keypoint-estimation}   & Emergency situations   \\
PHOENIX-2014T~\cite{2018_Neural_Sign_Language_Translation}                           & 9.2       & 0.6      & 0.7      & 2                         & 0.9                     & 1                        & 25.59~\cite{Vskou-2021-stochastic-transformer-SLT}   & Weather Forecast       \\
CSL Daily~\cite{Zhou2021ImprovingSLT-with-monolingual-CSLDaily}                      & 20.62     & 1.24     & 1.41     & 2                         & 1.3                     & 1.3                      & 23.92~\cite{chen2022simple} & Daily life             \\
OpenASL~\cite{shi-etal-2022-openASL}                            & $\leftarrow$ & 288  & $\rightarrow$          & $\leftarrow$ & 33 &$\rightarrow$                                                         & 6.72~\cite{shi-etal-2022-openASL}                          & Youtube (news + vlogs) \\ \midrule
How2Sign~\cite{Duarte_2021_how2sign}                          & 69.6      & 3.9      & 5.6      & 15.6                      & 3.2                     & 3.6                      & 8.03 (Ours)                                                                      & Instructional          \\ \bottomrule
\end{tabular}
\caption{Comparison between SLT datasets based on the duration of the videos (in hours), number of unique words (in thousands) in the vocabulary and SOTA on SLT without glosses. $\leftarrow$ $\rightarrow$ indicate that in some cases only statistics on the whole dataset are provided.}
\label{tab:SLT_datasets}
\end{table*}

Our baselines are similar to those published with OpenASL~\cite{shi-etal-2022-openASL}, another dataset of similar complexity. While the scores are not directly comparable because they are different datasets, the BLEU score of 6.72 reported for OpenASL is in a similar range to the values  we report in Section \ref{sec:results}.
Other works on alternative datasets of large scale obtained very poor BLEU scores: 1.0 in BOBSL~\cite{Albanie-2021-BOBSL}, 0.4 in SWISSTXT-NEWS~\cite{Camgoz-2021-content4all}, 0.4 in VRT-NEWS~\cite{Camgoz-2021-content4all}, or 0.37 in SRF~\cite{tarres-etal-2022-tackling} and 0.84 in FocusNews~\cite{microsoft-wmt-slt22} in the WMT shared task on sign language translation 2022~\cite{findings-wmt-slt22}.

\vspace{0.5em}
\noindent \textbf{Algorithms}.
SLT was initially approached with rule-based systems\cite{zhao-etal-2000-machine} and statistical methods\cite{Bungeroth2004StatisticalSL}. 
Since 2018, virtually all related work has basically applied the advances in deep learning to sign language translation datasets.
Given that SLT can be formulated as an input sequence of video frames that is transformed into a sequence of words, it fits perfectly in the popular sequence-to-sequence (seq2seq)~\cite{seq2seq-RNN} formulation widely adopted by the Machine Translation field which employs an encoder-decoder architecture to transform the input sequence into the output one, as depicted in Figure \ref{fig:teaser}.

First approaches in neural SLT used Recurrent Neural Networks (RNNs)~\cite{seq2seq-RNN} for the encoder-decoder architecture, whether with (GRUs) or LSTMs~\cite{Ko-2019-SLT-based-human-keypoint-estimation, Orbay-2020-SLT-learning-tokenization, Kim-2022-Keypointbased-SLT-without-glosses, Guo-2018-LSTM-SLT}.

However, RNNs present limitations in modeling long-term dependencies, an especially relevant problem when considering video input sequences captured at high frame rates.
To overcome these limitations, attention-based approaches were proposed~\cite{NMT-bahdanau}. Attention mechanisms selectively focus on parts of the input during decoding, allowing to capture long-term dependencies more effectively.
Camgoz et al. fed 2D CNN visual features into an RNN encoder-decoder with attention to perform translation~\cite{2018_Neural_Sign_Language_Translation}. 

The Transformer~\cite{vaswani2017attention} has emerged as the preferred option for numerous Natural Language Processing (NLP) and, more recently, Computer Vision (CV) tasks. The Transformer relies on the self-attention mechanism, which allows it to process the input sequence in parallel rather than sequentially, allowing even better modeling for long-term dependencies and allowing parallelization during training. 
Transformers have proven to work well for the SLT task~\cite{camgoz2020sign_language_transformers, yin-read-2020-better-sign-language-translation, Zhou2021ImprovingSLT-with-monolingual-CSLDaily, Chen-2022-twostream-SLR-SLT, camgoz2020multi, Camgoz-2021-content4all, li-2020-tspnet, Vskou-2021-stochastic-transformer-SLT}. Thus, we explore the Transformer architecture for the How2Sign dataset.

\vspace{0.5em}
\noindent \textbf{Tokenization of sign language videos}.
Sign language videos are normally tokenized to be fed into neural architectures, like the Transformer.
Camgoz et al.~\cite{2018_Neural_Sign_Language_Translation, camgoz2020sign_language_transformers} used 2D CNN to extract features of frames at the gloss-level.
These 2D CNN features were learned from another sign language recognition model~\cite{Zhou2021ImprovingSLT-with-monolingual-CSLDaily}.

Another commonly used features are inflated 3D convnets (I3D), developed for action-recognition~\cite{carreira-2017-i3d}, that can be further trained with sign language data~\cite{shi-etal-2022-openASL, albanie20_bsl1k, duarte2021signretrieval, li-2020-tspnet, microsoft-wmt-slt22,Orbay-2020-SLT-learning-tokenization}.
Similarly, \cite{chen2022simple} uses S3D \cite{Xie-2017-S3D} features, which have been pretrained in kinetics\cite{kay2017kinetics} and WLASL dataset. 
We focus our study on I3D video features.

Other works have also used pose estimators~\cite{OpenPose, MediaPipe} to represent the input video sequence since they contain relevant information about the motion and position of body parts. They have been particularly useful for action recognition~\cite{Shih-En-2016-pose-machines}.
Poses can be directly extracted, normalized and concatenated to form a video-level representation \cite{Ko-2019-SLT-based-human-keypoint-estimation, Camgoz-2021-content4all, findings-wmt-slt22} or perform frame level processing~\cite{Kim-2022-Keypointbased-SLT-without-glosses}.
Our work does not include baselines based on poses because our first efforts to train models using this approach were unsuccessful.

Finally, other manually designed and sophisticated multi-cue channels have been proposed for SLT.
In \cite{Chen-2022-twostream-SLR-SLT} they combine raw frames and poses with a two-stream network.
\cite{yin-read-2020-better-sign-language-translation} proposes using Spatial and Temporal Multi-cue networks~\cite{Zhou-2020-STMC}, which combines cues from image and pose (hand, face, full-frame and pose) in multiple scales. Another work uses a combination of raw frames and poses, and uses Graph Convolutional Networks to extract the tokens~\cite{Gan-2021-Skeleton-aware-SLT}.

\section{Data Preprocessing}
\label{sec:preprocessing}

One of the main challenges in SLT is the variability and complexity of sign languages, which can be influenced by a variety of factors such as the signer's background, context, and appearance. Therefore, it is important to preprocess the data to reduce this variability. This includes techniques such as visual feature extraction and normalization, as well as standardizing the format of the target data, which is text in our case.

\subsection{Video tokenization}
We choose I3D features~\cite{carreira-2017-i3d} to extract video representations directly from the RGB frames, motivated by their effectiveness in the sign recognition~\cite{MsASL, WLASL} and retrieval~\cite{duarte2021signretrieval} tasks. I3D features consider not only visual cues, but also temporal information. As a result, they provide a dense and reliable source of visual cues as input to our models.

The original I3D network is trained on ImageNet~\cite{imagenet} and fine-tuned for action recognition with the Kinetics-400~\cite{kay2017kinetics} dataset. As shown in~\cite{shi-etal-2022-openASL, albanie20_bsl1k, duarte2021signretrieval, li-2020-tspnet, microsoft-wmt-slt22,Orbay-2020-SLT-learning-tokenization}, further fine-tuning with sign language data is needed to properly model the temporal and spatial information present in them. We used the I3D features provided in~\cite{duarte2021signretrieval}.

The I3D network was trained on 16 consecutive frames, and videos were resized to 224 x 224. Color, scale, and horizontal flip augmentations were applied, and the features were extracted from the 1024-dimension activation before the pooling layer of the I3D backbone. Since they are already an output of a trained network, further processing, such as normalization, is not needed.

\subsection{Text processing}
\label{ssec:TextProcessing}
Text preprocessing is an important step in preparing raw text data into a more suitable format for NLP models. By cleaning, normalizing, and transforming text data into a standardized format, data can be effectively utilized by NLP algorithms.

\vspace{0.25em}
\noindent \textbf{Lowercase.} Similar to NLP pipelines, our system first converts raw text to lowercase. Lowercasing reduces the complexity of the vocabulary and minimizes the impact of irrelevant capitalization variations, thereby simplifying subsequent processing steps.

\vspace{0.25em}
\noindent \textbf{Tokenization.} We employ the Sentencepiece tokenizer~\cite{Kudo2018SentencePieceAS} to segment the lowercase text into sub-word units.
This approach represents a significant improvement over the conventional method of treating each word as a unit of the sequence. Word-based tokenization leads to an expansive vocabulary and an inability to account for previously unseen words, even if they are variations of words in the vocabulary.
On the other hand, sub-word tokenizers optimize the representation of words in the training data partition by identifying the most effective sub-word units, based on their frequency, while imposing a predefined vocabulary size. This allows better handling of unseen words, that can be represented as combinations of sub-words from the vocabulary.
Sub-word tokenization requires specifying a fixed vocabulary size, which becomes a hyperparameter to be optimized. The choice of vocabulary size has trade-offs in terms of representation and computational efficiency. When the vocabulary size is small, all sub-words are used more frequently, potentially leading to a better representation of unseen words. However, this also results in longer sequences as more sub-words are required to represent the same inputs, which can increase computational costs. Conversely, a larger vocabulary size reduces sequence length but may have worse coverage of rare and unseen words. Therefore, selecting the optimal vocabulary size requires balancing the need for better representation against the computational cost.

\vspace{0.25em}
\noindent \textbf{Postprocessing.} To ensure a fair assessment of the system's performance, it is necessary to compare the  model outputs to the original test set without any prior processing. However, this approach may result in a lower BLEU score, as the model generates text based on preprocessed data. For instance, comparing two versions of the same sentence, one lowercase and the other not, would result in the same word being counted as two different words. Therefore, we implement a postprocessing step, that involves detokenization and truecasing~\cite{lita-etal-2003-truecasing}, to restore the original capitalization and prevent this issue from arising.

\section{Methodology}
\label{sec:system}
\begin{figure}[h]
\centering
\includegraphics[width=1\linewidth]{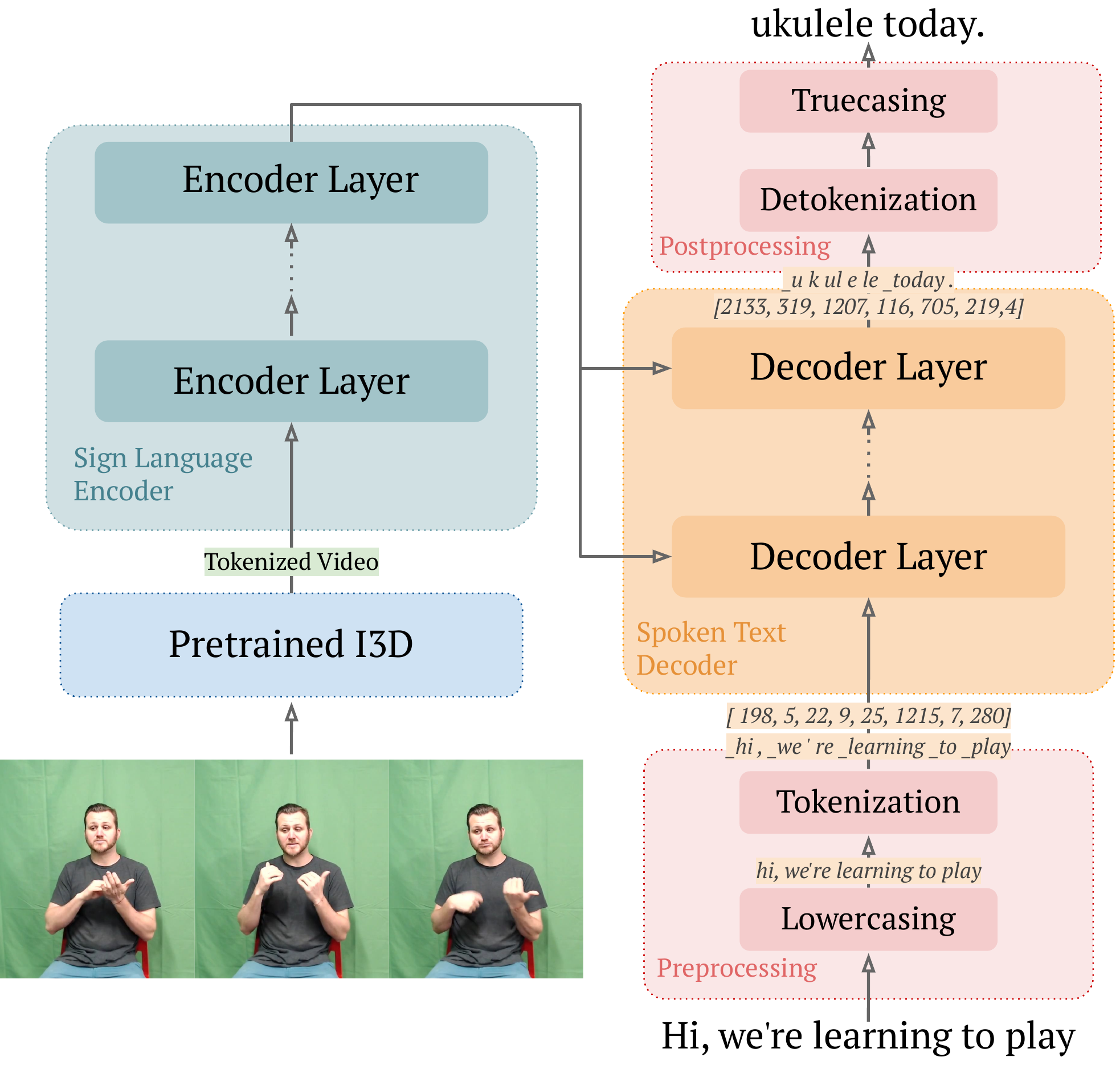}
\caption{The input video sequence is fed into a Transformer to generate the output text sequence.}
\label{fig:architecture}
\end{figure}

The building blocks of our implementation are depicted in Figure~\ref{fig:architecture}. 
The input video stream is tokenized with a pre-trained I3D feature extractor.
These tokens are fed into the encoding layers of the Transformer.
The decoder of the Transformer operates with lowercase and tokenized textual representations.

\subsection{Neural architecture}
We use a standard transformer encoder-decoder, to leverage their ability to model context and dependencies across the input sequence, as well as their sequence-to-sequence capabilities. 

For the model, we choose an asymmetric encoder-decoder with six encoder layers and three decoder layers, each with four attention heads, we select an embedding dimension of 256 and feed-forward network hidden size of 1024, which corresponds to ID (17) from Table~\ref{tab:ablation_parameters}.

\subsection{Implementation details}
In our implementation, we first preprocess the vocabulary as described in Section \ref{ssec:TextProcessing}, with a vocabulary size of $7000$ subwords.

For training, the batch size is set to 32, and we use cross entropy loss with label smoothing of $0.1$. We select the Adam optimizer, we warm-up the learning rate for the first $2000$ updates, and then we apply a cosine decay from $10^{-3}$ to $10^{-7}$ with warm restart every $1.7\cdot10^{4}$ steps
We train the model for $10^{5}$ steps, equivalent to $108$ epochs. We perform validation every two epochs.
Our training process takes 3.5 hours on a single NVIDIA GeForce RTX 2080 Ti GPU.

For inference, we adopt steps commonly used in machine translation.
During the text generation phase, the decoder predicts the next token by sampling from the probability distribution conditioned on the previously generated tokens. Instead of only selecting the prediction with higher probability, we use the beam search algorithm to generate predictions. Beam search generates multiple candidate sequences, we choose a beam size of five.

\begin{table*}[ht]
\centering
\begin{tabular}{ccccccccccc}
\toprule[0.5pt]
      & \multicolumn{5}{c}{\textbf{val}}        & \multicolumn{5}{c}{\textbf{test}}       \\
      & \textbf{rBLEU} & BLEU-1 & BLEU-2 & BLEU-3 & \textbf{BLEU} & \textbf{rBLEU} & BLEU-1 & BLEU-2 & BLEU-3 & \textbf{BLEU} \\ \midrule[0.3pt]
Ours. & 2.79  & 35.2   & 20.62 & 13.25  & 8.89 & 2.21  & 34.01  & 19.3  & 12.18  & \textbf{8.03} \\ \toprule[0.5pt]
\end{tabular}
\caption{Best scores on How2Sign for Sign Language Translation. }
\label{tab:SLT_results}
\end{table*}


\subsection{Evaluation protocol}
\label{sec:evaluation_protocol}
To measure the performance of our SLT models, we use BLEU score~\cite{papineni-etal-2002-bleu} a widely used metric in machine translation that measures the similarity between the predicted translation and the ground truth, at the corpus level.\footnote{Other SLT papers use BLEU-4 instead of BLEU. It represents the same score, we use BLEU for simplicity.} We implement it using sacreBLEU~\cite{post-2018-sacrebleu}.

The difficulty of the SLT task causes a bias in the model prediction towards most statistically frequent patterns, such as Example (2) and (3) in Table~\ref{tab:SLT_examples}.
These patterns can inflate the BLEU scores without actually translating anything meaningful.
Inspired by \cite{microsoft-wmt-slt22} we compute reducedBLEU (rBLEU).
This metric consists of removing certain words from the reference and the prediction before computing the BLEU score. We create a blacklist of words that are frequently used in the training data but do not contribute much to the meaning of the sentences, such as articles, prepositions, and pronouns.  
Appendix~\ref{appendix:blacklist} provides the complete list of the removed words, as well as the process to obtain them.

Table~\ref{tab:SLT_examples} shows a comparison between rBLEU and BLEU metrics. In general, rBLEU scores are substantially lower than BLEU. With rBLEU we are reducing the number of words of the sentences. Although How2Sign sentences are long, with an average of 11 words per sentence~\cite{Duarte_2021_how2sign}, if the reduced texts have less than four words, it is not possible to compute BLEU score and becomes zero-valued, even if the words are perfectly matching. 
For reference, for the test partition, after applying the reduction to the text, 40\% of the sentences contained less than 3 words.
Although only longer sentences contribute to the rBLEU metric, shorter sentences are known to be comparatively easier for the model, so we are able to identify models that deal better with complex examples.

Focusing on concrete examples, row (2) in Table~\ref{tab:SLT_examples}, shows that both the prediction and the reference contain the phrase ``In this clip I'm going to show you how to", which is one of the frequent patterns on the instructional dataset. This pattern inflates the BLEU score, while it does not affect the rBLEU score, which is low, suggesting that sentences have different meanings. 
Similarly, row (3) in Table~\ref{tab:SLT_examples} contains the phrase ``we're going to talk about how to", which similarly inflates the BLEU score even if prediction and the reference differ in meaning.

Experimental results indicate that rBLEU is a more reflective indicator of actual performance than traditional BLEU, for low-resource settings that also have repetitive patterns, given that it considers mostly semantically meaningful words. 
In order to provide comparable results with other works, we also report standard BLEU in our results.

\section{Experiments}
\label{sec:results}

The performance of our proposed approach is shown in Table~\ref{tab:SLT_results}. We evaluate our models using the metrics described in Section~\ref{sec:evaluation_protocol} and provide examples of generated spoken language translation sentences.

\subsection{Quantitative results}
Our implementation achieves the machine translation metrics reported in Table~\ref{tab:SLT_results}. To the authors' knowledge, these are the first published results for SLT obtained with the How2Sign dataset. 
The table displays the results of our best configuration, which provides a baseline from where future work can build upon.

\subsection{Qualitative results}
\label{sec:qualitative}
We provide a qualitative assessment of the results in Table~\ref{tab:SLT_examples}, showing a few spoken language translations generated by our best-performing model. Words used to compute rBLEU are in bold.

Example (1) shows the ability of our model to provide detailed translations even for complex words like \textit{``self defense"}. Our metrics indicate both high BLEU and rBLEU scores meaning that the model is generating a good translation, considering both full sentences and meaningful words.

However, our results also suggest that this is not always the case. For instance, in Examples (2) \& (3), BLEU is higher than rBLEU. We believe that this occurs because of the nature of the dataset. Given that we are working with instructional videos, there are frequent phrases like \textit{``I’m going to show you how to", ``we’re going to talk about how to"}, which the model learns to predict easily. And although this phrase has been correctly translated, the example has a different meaning, resulting in a lower rBLEU score.
Given that high BLEU scores can be misleading due to their susceptibility to frequent phrases, we emphasize the importance of using rBLEU instead of BLEU when selecting the best checkpoint.

The provided examples suggest that the models' performance may depend on the complexity and length of the signed video. We observed that the model was able to provide reasonably accurate translations for short sentences, except for Example (5). For longer sequences, the model struggled to capture the meaning of the video. This is also evidenced by the fact that only a few words are selected to compute rBLEU. 

The last example (6) illustrates the reason behind the disparity between rBLEU and BLEU metrics, explained in Section~\ref{sec:evaluation_protocol}. In this case, despite obtaining a high BLEU score and an accurate translation, the corresponding rBLEU score is zero due to the reduced number of remaining words for rBLEU calculation, which is less than four.

Overall, the findings suggest that the model's quality is still suboptimal, as demonstrated by Example (4), which has comparable metrics to the overall performance. 
Our analysis identifies cases where BLEU-guided translations fall short, and we propose rBLEU as a validation metric that aligns more closely with translations that effectively capture the original semantic meaning.

\begin{table}[b]
\centering
    \begin{tabular}{ll}
    \toprule
                               & Values                 \\ \midrule
    Text preprocessing         & \{\textbf{yes}, no\}            \\
    Vocabulary size            & \{1k, 4k, \textbf{7k}\}         \\ \midrule
    Batch size                 & \{\underline{\textbf{32}}, 64\}             \\
    Learning Rate (LR)         & \{5e-2, \underline{\textbf{1e-3}}, 5e-3\}   \\
    LR scheduler               & \{\textbf{cosine}, \underline{inv\_sqrt}\}  \\
    Warm-up steps              & \{0, \underline{\textbf{2k}}, 4k\}             \\
    Warm restarts period       & \{\underline{0},\textbf{17k}, 22k\}         \\
    Weight Decay               & \{\underline{1e-3}, 1e-2, \textbf{1e-1}\}            \\
    Label Smoothing            & \{\underline{0}, \textbf{0.1}\}             \\
    Dropout                    & \{0, \underline{0.1}, 0.2, \textbf{0.3}\}   \\
    \# Layers (encoder-decoder) & \{2-2, \underline{3-3}, 4-2, \textbf{6-3}\} \\
    Embed dim                  & \{\textbf{256}, \underline{512}\}           \\
    FFN dim                    & \{512, \textbf{1024}, \underline{2048}\}    \\
    \# Attention heads         & \{\textbf{4}, \underline{8}\}               \\ \bottomrule
    \end{tabular}
\caption{Hyperparameters search space. In bold are the optimal ones that we found during validation, and underlined are defaults.}
\label{tab:ablation_parameters}
\end{table}

\begin{table*}
\centering
\setlength{\tabcolsep}{12pt}
\vspace{-0.3cm}
\begin{tabular}{lllcc}
\multicolumn{2}{l}{\small Example}   &                                                                                                                                                                                                                                                     & rBLEU & BLEU \\ 
\hline
\multicolumn{1}{|l}{\multirow{2}{*}{(1)}} & \multicolumn{1}{l}{Reference}  &  \begin{tabular}[c]{@{}l@{}}And that's a \textbf{great} \textbf{vital} \textbf{point} \textbf{technique} for \textbf{women's} \textbf{self} \textbf{defense}.\end{tabular}                                                                                                                                                                                                                                         & \multirow{2}{*}{30.29}                    & \multicolumn{1}{c|}{\multirow{2}{*}{38.25}} \\
\multicolumn{1}{|l}{}                     & \multicolumn{1}{l}{Prediction}  & It's really a \textbf{great} \textbf{point} for \textbf{women's} \textbf{self} \textbf{defense}.                                                                                                                                                                                                                                                                                                   &                                           & \multicolumn{1}{c|}{}                       \\ \hline
\multicolumn{1}{|l}{\multirow{2}{*}{(2)}} & \multicolumn{1}{l}{Reference}  & In this \textbf{clip} I'm going to \textbf{show} you how to \textbf{tape} your \textbf{cables} down.      & \multirow{2}{*}{24.88}   & \multicolumn{1}{c|}{\multirow{2}{*}{64.53}} \\
\multicolumn{1}{|l}{}                     & \multicolumn{1}{l}{Prediction}  & In this \textbf{clip} I'm going to \textbf{show} you how to \textbf{improve} \textbf{push} \textbf{ups}.                &                         & \multicolumn{1}{c|}{}    \\ \hline
\multicolumn{1}{|l}{\multirow{2}{*}{(3)}} & \multicolumn{1}{l}{Reference}  & \begin{tabular}[c]{@{}l@{}}In this \textbf{segment} we're going to \textbf{talk} about how to \textbf{load} your still for \\\textbf{distillation} of \textbf{lavender} \textbf{essential} \textbf{oil}.\\ -----\end{tabular}                                                                                                                                                                                               & \multirow{2}{*}{6.77}                     & \multicolumn{1}{c|}{\multirow{2}{*}{29.82}} \\
\multicolumn{1}{|l}{}                     & \multicolumn{1}{l}{Ours}  & \begin{tabular}[c]{@{}l@{}}\textbf{Ok}, in this \textbf{clip}, we're going to \textbf{talk} about how to \textbf{fold} the \textbf{ink} for \\ the \textbf{lid} of the \textbf{oil}.\end{tabular}                                                                                                                                                                                                                    &                                           & \multicolumn{1}{c|}{}                       \\ \hline
\multicolumn{1}{|l}{\multirow{2}{*}{(4)}} & \multicolumn{1}{l}{Reference}  & \begin{tabular}[c]{@{}l@{}}You are \textbf{dancing}, and now you are going to \textbf{need} the \textbf{veil} and you \\ are going to just \textbf{grab} the \textbf{veil} as \textbf{far} as \textbf{possible}.\\ -----\end{tabular}                                                                                                                                                                                        & \multicolumn{1}{c}{\multirow{2}{*}{4.93}} & \multicolumn{1}{c|}{\multirow{2}{*}{8.04}}  \\
\multicolumn{1}{|l}{}                     & \multicolumn{1}{l}{Ours.} & \begin{tabular}[c]{@{}l@{}}So, \textbf{once} you're \textbf{belly} \textbf{dancing}, \textbf{once} you've \textbf{got} to have the \textbf{strap}, \\ you're going to \textbf{need} to \textbf{grab} the \textbf{thumb}, and \textbf{try} to \textbf{avoid} it.\end{tabular}                                                                                                                                                                             & \multicolumn{1}{l}{}                      & \multicolumn{1}{l|}{}                       \\ \hline
\multicolumn{1}{|l}{\multirow{2}{*}{(5)}} & \multicolumn{1}{l}{Reference}  & \begin{tabular}[c]{@{}l@{}}But if you have to \textbf{setup} a \textbf{new campfire}, \textbf{there's two} \textbf{ways} to do \\ it in a very \textbf{low impact}; one is with a \textbf{mound fire}, which we should \\in the \textbf{campfire segment earlier} and the other way to \textbf{setup} a \textbf{low}\\ \textbf{impact campfire} is to have a \textbf{fire pan}, which is just a \textbf{steel pan} like \\ the \textbf{top} of a \textbf{trash} can.\\ -----\end{tabular} & \multirow{2}{*}{0.85}                     & \multicolumn{1}{c|}{\multirow{2}{*}{3.79}}  \\
\multicolumn{1}{|l}{}                     & \multicolumn{1}{l}{Ours}  & \begin{tabular}[c]{@{}l@{}}And other thing I'm going to \textbf{talk} to you is a little bit more \textbf{space}, \\ a \textbf{space} that's what it's going to do, it's \textbf{kind} of a \textbf{quick}, and then I \\don't want to take a \textbf{spray skirt} off, and then I don't want it to take\\ it to the \textbf{top} of it.\end{tabular}                                                              &                                           & \multicolumn{1}{c|}{}                       \\ \hline
\multicolumn{1}{|l}{\multirow{2}{*}{(6)}} & \multicolumn{1}{l}{Reference}  & \begin{tabular}[c]{@{}l@{}}So, this is a very \textbf{important} \textbf{part} of the \textbf{process}.\end{tabular}                                                                                  & \multirow{2}{*}{0.0}                        & \multicolumn{1}{c|}{\multirow{2}{*}{61.86}} \\
\multicolumn{1}{|l}{}                     & \multicolumn{1}{l}{Ours}  & It's a very \textbf{important} \textbf{part} of the \textbf{process}.                                                                                                                                                                                                                                                                                                            &                                           & \multicolumn{1}{c|}{}                       \\ \hline
\end{tabular}
\caption{Qualitative examples from our best-performing model. In bold the words remaining to compute rBLEU. Corresponding manually selected input frames from examples can be found in Appendix~\ref{appendix:examples_frames}}
\label{tab:SLT_examples}
\end{table*}

\subsection{Hyperparameter search}
Transformer under low-resource conditions is highly dependent on hyperparameter settings~\cite{araabi-monz-2020-optimizing}. Our experiments show that using an optimized Transformer improves the translation quality over 3.47 BLEU points and 1.8 reduced BLEU points compared to the default hyperparameters for SLT.

Table~\ref{tab:ablation_parameters} shows the hyperparameters that we optimize, ordered by tuning order. 
Default hyperparameters for SLT come from~\cite{camgoz2020sign_language_transformers}.
Exploring all possible values in Table~\ref{tab:ablation_parameters} is extremely expensive. Possible methodologies of exploration include random search or grid search. We choose a flexible grid search, which means that we try different values, and once the hyperparameter is tuned, we fix it. Since there is no guarantee that this will result in a global optimum, we analyze the results to discard or add experiments during the exploration.

As highlighted in Section~\ref{sec:preprocessing}, text preprocessing plays an important role in NLP tasks. In our experiments, we opted to lowercase our training data. To showcase its efficacy, we trained our model with both preprocessed text data and raw text data (i.e., direct production of truecased outputs). Our results indicate that lowercasing the text data yields significant improvements in the rBLEU metric, as evidenced in Table~\ref{tab:Ablation_preprocessing}.

\vspace{1em}

\begin{table}[hbt!]
\centering
\begin{tabular}{ccc}
\toprule
ID  & \begin{tabular}[c]{@{}c@{}}Text\\ Preprocessing\end{tabular} & rBLEU \\ \midrule
(1) & no                                                           & 0.62       \\
(2) & yes                                                          & \textbf{0.98}       \\ \bottomrule
\end{tabular}
\caption{Impact of text preproceessing.}
\label{tab:Ablation_preprocessing}
\end{table}
\begin{table}[hbt!]
\centering
\begin{tabular}{ccc}
\toprule
ID  & \multicolumn{1}{c}{\begin{tabular}[c]{@{}c@{}}Vocabulary\\ Size\end{tabular}} & rBLEU \\ \midrule
(3) & 1000                                                                            & 0.85       \\
(4) & 4000                                                                            & 0.89       \\
(5) & 7000                                                                            & \textbf{0.98}       \\ \bottomrule
\end{tabular}
\caption{Impact of the vocabulary size.}
\label{tab:ablation_dict}
\end{table}

\begin{table*}[hbt!]
\centering
\begin{tabular}{ccccccccc}
\toprule
ID   & \begin{tabular}[c]{@{}c@{}}Encoder\\ Layers\end{tabular} & \begin{tabular}[c]{@{}c@{}}Decoder\\ Layers\end{tabular} & \begin{tabular}[c]{@{}c@{}}Embed\\ Dim\end{tabular} & \begin{tabular}[c]{@{}c@{}}FFN\\ Dim\end{tabular} & \begin{tabular}[c]{@{}c@{}}Attention\\ Heads\end{tabular} & LR    & \begin{tabular}[c]{@{}c@{}}LR\\ Scheduler\end{tabular} & rBLEU \\ \midrule

(6)     & 3   & 3   & 512   & 2048  & 8     & 0.001     & inverse\_sqrt   & 0.98    \\
(7)     & 3   & 3   & 512   & 2048  & 8     & 0.001     & cosine (T=17k)  & 0.89    \\
(8)     & 3   & 3   & 256   & 1024  & 4     & 0.001     & cosine (T=17k)  & 1.14    \\
(9)     & 3   & 3   & 256   & 1024  & 4     & 0.005     & cosine (T=17k)  & 0.68    \\
\midrule
(10)    & 2   & 2   & 256   & 1024  & 4     & 0.001     & cosine (T=17k)  & 1.32    \\
(11)    & 2   & 2   & 256   & 1024  & 4     & 0.005     & cosine (T=17k)  & 0.72    \\
(12)    & 2   & 2   & 256   & 512   & 4     & 0.001     & cosine (T=17k)  & 1.37    \\
\midrule
(13)    & 4   & 2   & 256   & 1024  & 4     & 0.001     & cosine (T=17k)  & 1.14    \\
(14)    & 4   & 2   & 256   & 1024  & 4     & 0.005     & cosine (T=17k)  & 0.64    \\
\midrule
(15)    & 6   & 3   & 512   & 2048  & 8     & 0.001     & cosine (T=17k)  & 0.87     \\
(16)    & 6   & 3   & 512   & 2048  & 8     & 0.001     & cosine (T=22k)  & 0.75        \\
(17)    & 6   & 3   & 256   & 1024  & 4     & 0.001     & cosine (T=17k)  & 0.93       \\ \bottomrule
\end{tabular}
\caption{Validation scores during the exploration of the model architecture.}
\label{tab:ablation_model}
\end{table*}

As previously discussed in Section~\ref{ssec:TextProcessing}, the selection of appropriate vocabulary size is a trade-off between enhancing the representation of rare words or producing shorter sequences. In our study, we utilize the SentencePiece~\cite{Kudo2018SentencePieceAS} tokenizer and experiment with dictionary sizes of $1000$, $4000$, and $7000$ sub-words to evaluate their respective impacts on NMT performance.

Results in Table~\ref{tab:ablation_dict} show that larger dictionary improves results, in our experiments it increases 0.14 points of reduced BLEU score. Thus, we always use a Sentencepiece tokenizer with a vocabulary of $7000$ sub-words.

A current observation in Transformers is that increasing the number of parameters will improve the performance. However, in low-resource languages, increasing the number of model parameters can hinder performance \cite{yin-read-2020-better-sign-language-translation}.
We study the effect of using a deeper and shallower Transformer by changing the number of layers in the encoder and decoder, the number of attention heads, the feed-forward layer dimension, and embedding dimensions.

Since the optimization of the learning rate (LR) is dependent on the number of parameters of the model, we tune it together with other hyperparameters related to the architecture size. Furthermore, we introduce the use of LR scheduling of cosine with warm restarts. This scheduler has been shown to perform better than alternatives~\cite{scheduler}. The resetting of the learning rate acts as a simulated restart of the learning process and is defined by number of steps T.

Table~\ref{tab:ablation_model} shows the results of our system optimizations. 
Experiments point to the direction that smaller models, like (12) perform better for our dataset. The loss curves indicated a substantial amount of overfitting in the larger models, which is most likely related to the \textit{small} amount of provided data compared to the amount of data needed to tune a large number of parameters.
We see gains in tuning the learning rate to improve performance.
Our results indicate that it is beneficial to use four attention heads instead of eight under low-resource conditions.
Due to the fact that the input data is by far more complex than the output, we choose to carry out further experiments with both the best symmetric model (12) and the best asymmetric model (17).

Given the observed overfitting, we add regularization by adding dropout, weight decay, and label smoothing. Considering it is difficult to perform data augmentation with our video features, adding regularization helps make the model more robust to overfitting.

Table~\ref{tab:ablation_regu} shows that we obtain substantial improvements by increasing regularization techniques. That is to be expected since overfitting was present in our previous experiments. Surprisingly, it appears that tuning these hyperparameters is the most effective measure to improve the model's performance.
We also show that under these conditions, a larger model paired, such as (22), with regularization techniques outperforms a smaller model.

\begin{table}[hb!]
\centering
    \begin{tabular}{ccccccc}
    \toprule
    ID & Base   & Dropout & \begin{tabular}[c]{@{}l@{}}Weight\\ Decay\end{tabular} & \multicolumn{1}{c}{\begin{tabular}[c]{@{}c@{}}Label\\Smoothing\end{tabular}} & rBLEU \\ \midrule
    (6) & - & 0.1     & 0.001                                                  & 0                                                                                & 0.98       \\
    (18) & (17) & 0.2     & 0.01                                                   & 0.1                                                                              & 1.17       \\
    
    (19) & (12) & 0.2     & 0.01                                                    & 0.1                                                                              & 1.21       \\ 
    (20) & (6) & 0.3     & 0.1                                                   & 0.1                                                                              & 1.84       \\
    (21) & (12) & 0.3     & 0.1                                                    & 0.1                                                                              & 1.38       \\ 
    (22) & (17) & 0.3     & 0.1                                                   & 0.1                                                                              & \textbf{2.78}       \\ 
    
    \bottomrule
    \end{tabular}
\caption{Validation scores for different regularization techniques.}
\label{tab:ablation_regu}
\end{table}

\section{Discussion}
\label{sec:discussion}
Our experiments yielded several findings. Firstly, we observed that text preprocessing is an important step that can significantly improve performance, resulting in an increase of 0.36 rBLEU points. Secondly, we found using a greater vocabulary size led to an increase of 0.14 rBLEU. 

Another finding was that choosing the correct parameters for the architecture is crucial for achieving optimal performance, resulting in a 0.39 rBLEU improvement. Furthermore, our results highlight the difficulty of finding the sweet spot where regularization techniques help but not hinder the performance of deep learning models. In our case, we boosted performance by an impressive 1.8 rBLEU points after an extensive sweep of hyperparameters and configurations.

During the qualitative analysis, we show that the model is able to produce meaningful translations. Moreover, our experiment highlights the importance of considering rBLEU as an effective metric for evaluating the best checkpoint. Higher rBLEU scores 
indicated a consistent correlation with the model's ability to capture the semantic meaning from the video.

While our work has shown promising results, there is still room for improvement. 
Our current approach only explores the use of I3D as visual feature. While other works use pose landmarks as a visual features~\cite{Ko-2019-SLT-based-human-keypoint-estimation, Camgoz-2021-content4all, findings-wmt-slt22, Kim-2022-Keypointbased-SLT-without-glosses}, 
our initial exploratory work with MediaPipe~\cite{MediaPipe} poses, obtained unsatisfactory results with BLEU score of 0.8 for the test partition. 

Furthermore, upon qualitative exploration, we realized the decoder was discarding the conditioning provided by the encoder and functioning solely as a language model. We hypothesize that this behavior may be due to our current approach of feeding poses as sequences of one-dimensional arrays containing only landmark coordinates. This method may not be the most effective way of processing the graph-like structure present in poses. One proposed way of tackling this is extracting optical flow features based on human pose estimation~\cite{SLT_pose_amit}, which worked well for sign language detection.
Similarly to \cite{yin-etal-2021-including-signed-languages}, we recognize the need for an in-depth exploration of visual features appropriate for SLT.

We believe another exciting direction would be the exploration of using a pre-trained decoder, similar to~\cite{de2021frozen}, where language models that are already trained for spoken language translation are adapted for sign languages.

\vspace{0.5em}
\noindent \textbf{Societal Impact.} Efficiently translating sign language videos can have a significant impact on accessibility, opening up a range of useful applications.
However, there are also potential risks associated with this technology, including problems associated with the accuracy of models, which currently produce inaccurate or incomplete translations, biases present in the datasets, and increased risk of surveillance of signers, similarly how automatic speech recognition (ASR) technologies may affect the privacy of speakers.

\section{Conclusions}
\label{sec:conclusions}
In this work, we made an open-source implementation that serves as a first baseline for sign language translation on the How2Sign dataset, a large and complex dataset. Our approach achieved a BLEU score of 8.03, inidicating a certain level of understanding of the signed utterances, which is on par with results reported for OpenASL~\cite{shi-etal-2022-openASL}, a publicly available dataset of comparable complexity.

Additionally, our extensive hyperparameter search demonstrates the necessity of tuning to obtain the best set of parameters. The best results are obtained with an asymmetric Transformer trained with great amounts of regularization.

Our evaluations, both quantitative and qualitative, have led us to conclude that rBLEU is a suitable evaluation metric for similar benchmarks, particularly for low-resource datasets with frequent repetitive patterns. In contrast to traditional BLEU score, which may be inflated due to these patterns, rBLEU provides a more accurate evaluation that better reflects the model's performance.

Lastly, we provide the code and models to allow reproducibility and encourage further research and advancements in sign language translation field.

\section*{Acknowledgements}
This research was partially supported by research grant Adavoice PID2019-107579RB-I00 / AEI / 10.13039/501100011033, research grants PRE2020-094223, PID2021-126248OB-I00 and PID2019-107255GB-C21 and by Generalitat de Catalunya (AGAUR) under grant agreement 2021-SGR-00478. 

{\small
\bibliographystyle{ieee_fullname}
\bibliography{references}
}

\appendix
\begin{appendices}
\onecolumn
\section{Appendix}
\subsection{Complete List of reduction words to compute rBLEU}
\label{appendix:blacklist}

We present the complete list of reduction words to compute rBLEU. We select the most frequent words in the How2Sign train partition. We manually clean the list to remove any words that contain meaning, mostly those that do not fall into the category of articles, prepositions, or pronouns.

\begin{multicols}{9}
the

to

and

you

a

of

that

is

it

in

your

going

this

so

on

i

have

can

with

are

just

we

be

want

for

do

if

or

up

it's

as

you're

like

get

one

i'm

now

what

little

out

then

make

here

go

but

not

about

we're

when

they

all

there

my

will

at

how

some

right

back

really

use

very

down

don't

that's

because

our

way

from

them

take

more

put

an

into

sure

bit

also

these

would

know

off

lot

over

time

which

thing

other

where

again

any

actually

has

come

there's

through

by

her

much

their

those

they're

he

might

me

you'll

could

you've

than

was

let's

should

we'll

i've

his

end

able

doesn't

many

every

we've

each

does

been

him

i'll

yourself

am

had

she

were

while

its

who

he's

why

already

either

us

whatever

did

she's

can't

cause

what's

\end{multicols}

\newpage

\subsection{Table~\ref{tab:SLT_examples} with corresponding input frames}
\label{appendix:examples_frames}
\vspace{-2em}
\begin{table*}[b]
\centering
\setlength{\tabcolsep}{12pt}
\vspace{-0.3cm}
\begin{tabular}{lllcc}
ID  &    &                                                                                                                                                                                                                                                     & rBLEU & BLEU \\ 
\hline
\multicolumn{1}{|l}{} & Input & \includegraphics[width=0.3\textwidth]{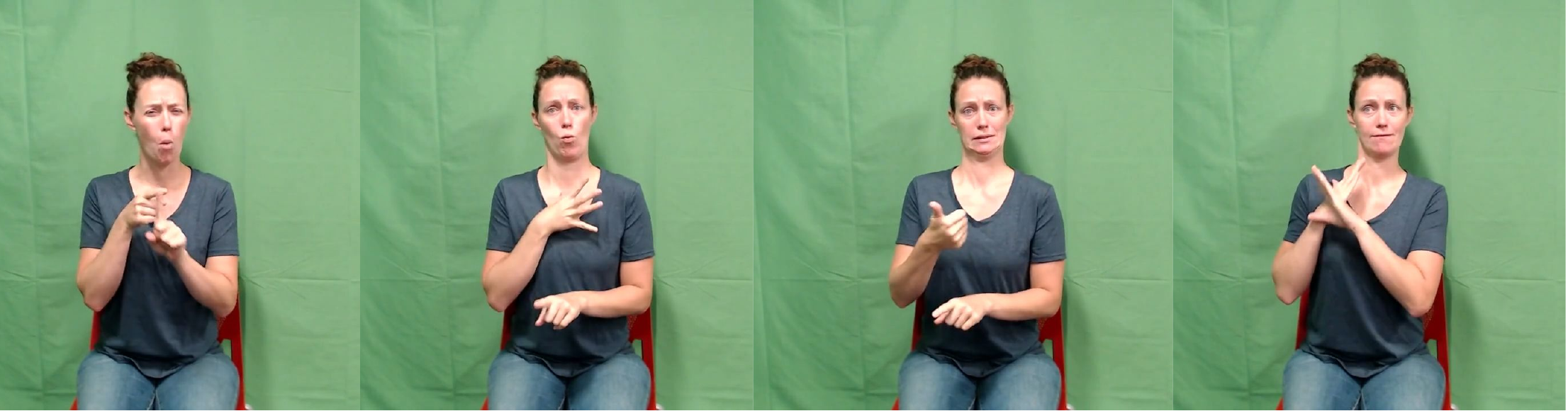} & & \multicolumn{1}{l|}{}\\
\multicolumn{1}{|l}{\multirow{2}{*}{(1)}} & \multicolumn{1}{l}{Reference}  &  \begin{tabular}[c]{@{}l@{}}And that's a \textbf{great} \textbf{vital} \textbf{point} \textbf{technique} for \textbf{women's} \textbf{self} \textbf{defense}.\end{tabular}                                                                                                                                                                                                                                         & \multirow{2}{*}{30.29}                    & \multicolumn{1}{c|}{\multirow{2}{*}{38.25}} \\
\multicolumn{1}{|l}{}                     & \multicolumn{1}{l}{Prediction}  & It's really a \textbf{great} \textbf{point} for \textbf{women's} \textbf{self} \textbf{defense}.                                                                                                                                                                                                                                                                                                   &                                           & \multicolumn{1}{c|}{}                       \\ \hline
\multicolumn{1}{|l}{} & Input & \includegraphics[width=0.3\textwidth]{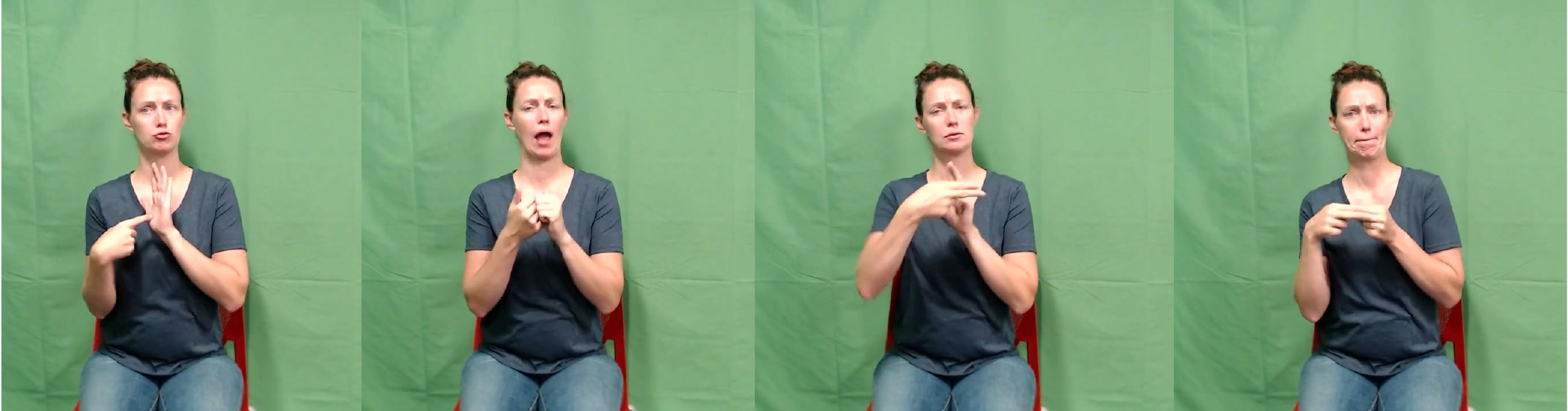} & & \multicolumn{1}{l|}{}\\
\multicolumn{1}{|l}{\multirow{2}{*}{(2)}} & \multicolumn{1}{l}{Reference}  & In this \textbf{clip} I'm going to \textbf{show} you how to \textbf{tape} your \textbf{cables} down.      & \multirow{2}{*}{24.88}   & \multicolumn{1}{c|}{\multirow{2}{*}{64.53}} \\
\multicolumn{1}{|l}{}                     & \multicolumn{1}{l}{Prediction}  & In this \textbf{clip} I'm going to \textbf{show} you how to \textbf{improve} \textbf{push} \textbf{ups}.                &                         & \multicolumn{1}{c|}{}    \\ \hline
\multicolumn{1}{|l}{} & Input & \includegraphics[width=0.3\textwidth]{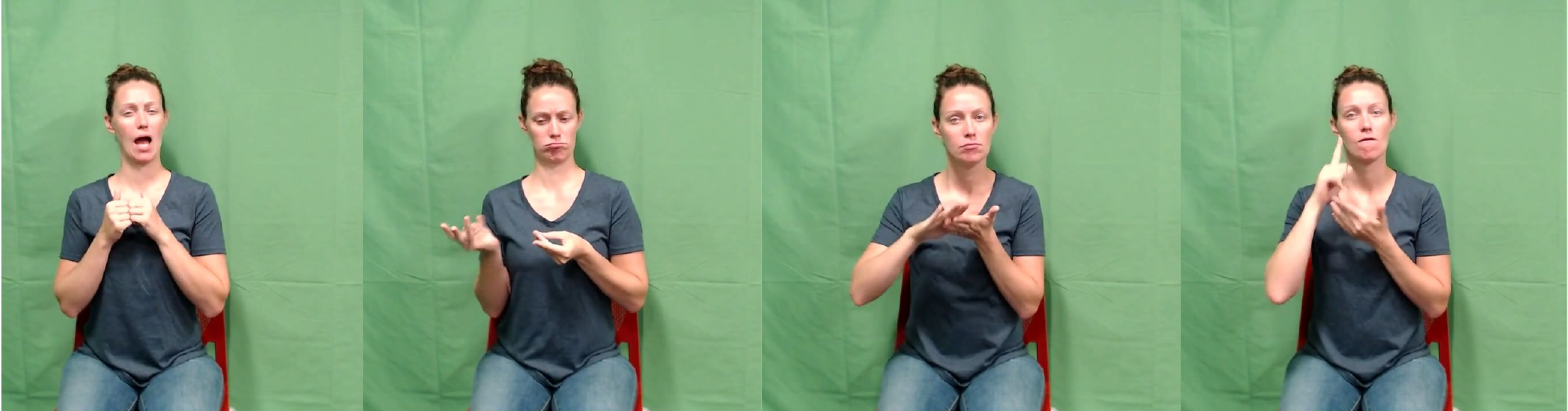} & & \multicolumn{1}{l|}{}\\
\multicolumn{1}{|l}{\multirow{2}{*}{(3)}} & \multicolumn{1}{l}{Reference}  & \begin{tabular}[c]{@{}l@{}}In this \textbf{segment} we're going to \textbf{talk} about how to \textbf{load} your still for \\\textbf{distillation} of \textbf{lavender} \textbf{essential} \textbf{oil}.\\ -----\end{tabular}                                                                                                                                                                                               & \multirow{2}{*}{6.77}                     & \multicolumn{1}{c|}{\multirow{2}{*}{29.82}} \\
\multicolumn{1}{|l}{}                     & \multicolumn{1}{l}{Ours}  & \begin{tabular}[c]{@{}l@{}}\textbf{Ok}, in this \textbf{clip}, we're going to \textbf{talk} about how to \textbf{fold} the \textbf{ink} for \\ the \textbf{lid} of the \textbf{oil}.\end{tabular}                                                                                                                                                                                                                    &                                           & \multicolumn{1}{c|}{}                       \\ \hline
\multicolumn{1}{|l}{} & Input & \includegraphics[width=0.3\textwidth]{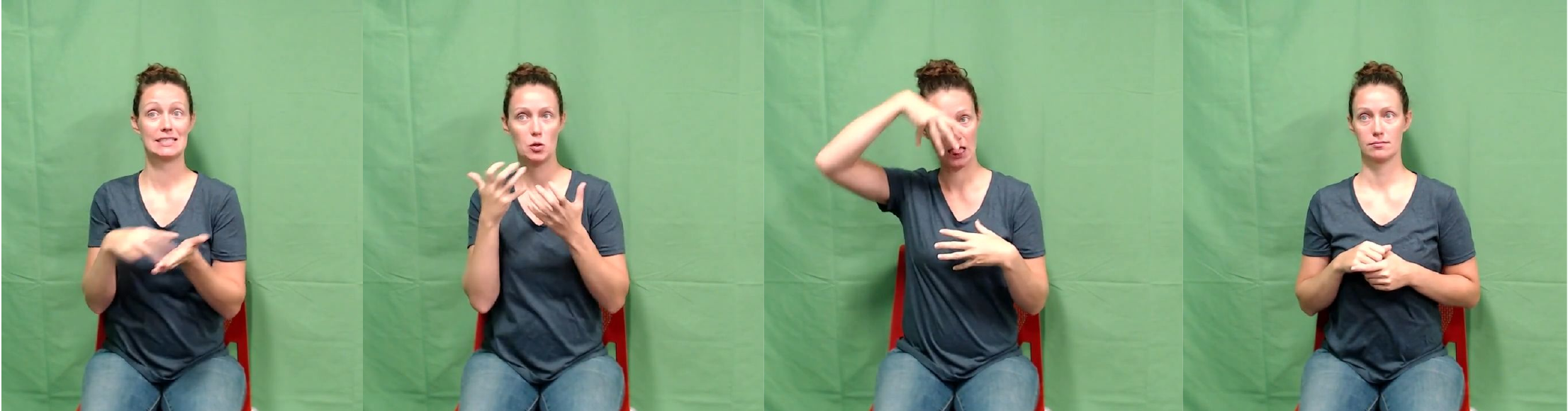} & & \multicolumn{1}{l|}{}\\
\multicolumn{1}{|l}{\multirow{2}{*}{(4)}} & \multicolumn{1}{l}{Reference}  & \begin{tabular}[c]{@{}l@{}}You are \textbf{dancing}, and now you are going to \textbf{need} the \textbf{veil} and you \\ are going to just \textbf{grab} the \textbf{veil} as \textbf{far} as \textbf{possible}.\\ -----\end{tabular}                                                                                                                                                                                        & \multicolumn{1}{c}{\multirow{2}{*}{4.93}} & \multicolumn{1}{c|}{\multirow{2}{*}{8.04}}  \\
\multicolumn{1}{|l}{}                     & \multicolumn{1}{l}{Ours.} & \begin{tabular}[c]{@{}l@{}}So, \textbf{once} you're \textbf{belly} \textbf{dancing}, \textbf{once} you've \textbf{got} to have the \textbf{strap}, \\ you're going to \textbf{need} to \textbf{grab} the \textbf{thumb}, and \textbf{try} to \textbf{avoid} it.\end{tabular}                                                                                                                                                                             & \multicolumn{1}{l}{}                      & \multicolumn{1}{l|}{}                       \\ \hline
\multicolumn{1}{|l}{} & Input & \includegraphics[width=0.3\textwidth]{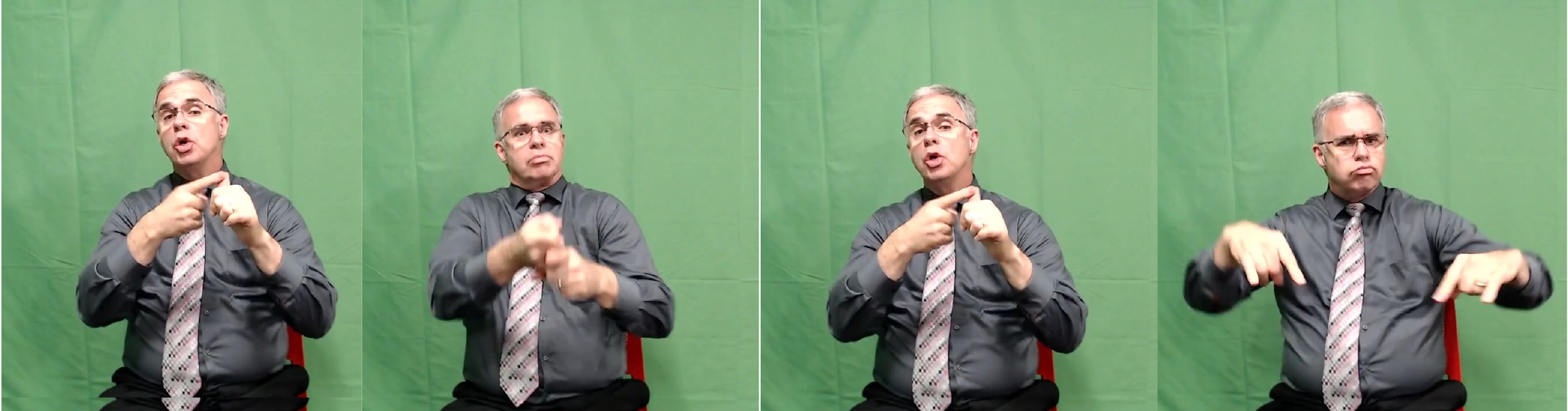} & & \multicolumn{1}{l|}{}\\
\multicolumn{1}{|l}{\multirow{2}{*}{(5)}} & \multicolumn{1}{l}{Reference}  & \begin{tabular}[c]{@{}l@{}}But if you have to \textbf{setup} a \textbf{new campfire}, \textbf{there's two} \textbf{ways} to do \\ it in a very \textbf{low impact}; one is with a \textbf{mound fire}, which we should \\in the \textbf{campfire segment earlier} and the other way to \textbf{setup} a \textbf{low}\\ \textbf{impact campfire} is to have a \textbf{fire pan}, which is just a \textbf{steel pan} like \\ the \textbf{top} of a \textbf{trash} can.\\ -----\end{tabular} & \multirow{2}{*}{0.85}                     & \multicolumn{1}{c|}{\multirow{2}{*}{3.79}}  \\
\multicolumn{1}{|l}{}                     & \multicolumn{1}{l}{Ours}  & \begin{tabular}[c]{@{}l@{}}And other thing I'm going to \textbf{talk} to you is a little bit more \textbf{space}, \\ a \textbf{space} that's what it's going to do, it's \textbf{kind} of a \textbf{quick}, and then I \\don't want to take a \textbf{spray skirt} off, and then I don't want it to take\\ it to the \textbf{top} of it.\end{tabular}                                                              &                                           & \multicolumn{1}{c|}{}                       \\ \hline
\multicolumn{1}{|l}{} & Input & \includegraphics[width=0.3\textwidth]{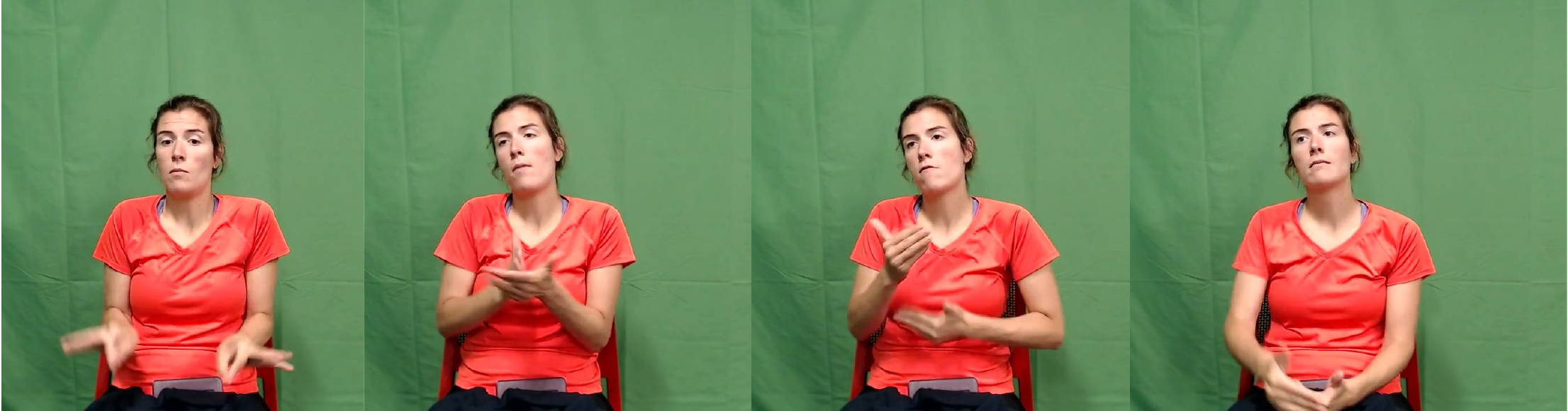} & & \multicolumn{1}{l|}{}\\
\multicolumn{1}{|l}{\multirow{2}{*}{(6)}} & \multicolumn{1}{l}{Reference}  & \begin{tabular}[c]{@{}l@{}}So, this is a very \textbf{important} \textbf{part} of the \textbf{process}.\end{tabular}                                                                                  & \multirow{2}{*}{0}                        & \multicolumn{1}{c|}{\multirow{2}{*}{61.86}} \\
\multicolumn{1}{|l}{}                     & \multicolumn{1}{l}{Ours}  & It's a very \textbf{important} \textbf{part} of the \textbf{process}.                                                                                                                                                                                                                                                                                                            &                                           & \multicolumn{1}{c|}{}                       \\ \hline
\end{tabular}
\caption{Qualitative examples from our best-performing model. In bold the words remaining to compute rBLEU. Together with selected frames from the input video.}
\label{tab:SLT_examples_images}
\end{table*}

\end{appendices}

\end{document}